# PhysSFI-Net: physics-informed geometric learning of skeletal and facial interactions for orthognathic surgical outcome prediction

Check for updates

**Jiahao Bao[1,2,3,8], Huazhen Liu[4,8], Yu Zhuang[1,2,3,8], Leran Tao[1,2,3,8], Xinyu Xu[4], Yongtao Shi[5], Mengjia Cheng[6], Yiming Wang[1,2,3], Congshuang Ku[1,2,3], Ting Zeng[1,2,3], Yilang Du[1,2,3], Siyi Chen[1,2,3], Shunyao Shen[1,2,3,9], Suncheng Xiang[7,9] & Hongbo Yu[1,2,3,9]** ✉

Orthognathic surgery repositions jaw bones to restore occlusion and enhance facial aesthetics. Accurate simulation of postoperative facial morphology is essential for preoperative planning. This study aims to develop and validate a physics-informed geometric deep learning framework named PhysSFI-Net for precise prediction of soft tissue deformation following orthognathic surgery. The model integrates a hierarchical feature extraction module with attention mechanisms to capture skeletal-facial interactions, an LSTM-based sequential predictor for incremental deformation, and a biomechanics-inspired reconstruction module for high-resolution facial modeling. The model was trained on 135 patients and externally validated on an independent cohort of 33 patients. Model performance was assessed using point cloud shape error, surface deviation error and landmark error between predicted facial shapes with corresponding ground truths. Quantitative analysis demonstrated that PhysSFI-Net achieved a global shape error of 1.070 ± 0.088 mm, a surface deviation error of 1.296 ± 0.349 mm and a landmark error of 2.445 ± 1.326 mm. Comparative experiments indicated that PhysSFI-Net outperformed the state-of-the-art method ACMT-Net and baseline models. External validation further confirmed its robustness with a global HD of 1.431 ± 0.087 mm and consistently lower subregional and mesh-based errors. In conclusion, PhysSFI-Net enables interpretable, high-resolution prediction of postoperative facial morphology, showing strong potential for clinical application in orthognathic surgical planning.

Dento-maxillofacial deformities, characterized by dysfunctions in the stomatognathic system and abnormal facial morphology, significantly impact patients' physiological functions and mental well-being[1]. Orthognathic surgery (OGS) corrects jaw deformities by performing osteotomies and repositioning bone segments of the maxilla and mandible, thereby restoring occlusal function and enhancing facial aesthetics[2,3]. The preoperative optimization of the surgical plan is a crucial determinant of successful outcomes in OGS[4]. Computer-Assisted Surgical Simulation (CASS) provides clinicians with powerful tools for virtual surgery and decision support. By reconstructing three-dimensional (3D) maxillofacial models, it enables comprehensive evaluation of anatomical structures and allows precise simulation of osteotomies and bony segment movements, thus facilitating the identification of optimal surgical strategies[5,6]. During the process of CASS, occlusion can be accurately simulated through dental model articulation due to the rigid contact relationship between teeth[7]. However, the deformation of facial soft tissues following skeletal movement remains challenging to precisely predict, owing to the nonlinear and complex biomechanical relationship between soft tissues and underlying bony structures[8]. This leads to surgical planning decisions that lack objective criteria and are highly reliant on clinical experience. Hence, there is an urgent need for developing an accurate and effective predictive approach.

[1]Department of Oral and Cranio-maxillofacial Surgery, Shanghai Ninth People's Hospital, College of Stomatology, Shanghai Jiao Tong University School of Medicine, Shanghai, China. [2]National Center for Stomatology & National Clinical Research Center for Oral Diseases, Shanghai, China. [3]Shanghai Key Laboratory of Stomatology & Shanghai Research Institute of Stomatology, Shanghai, China. [4]School of Electronic Information and Electrical Engineering, Shanghai Jiao Tong University, Shanghai, China. [5]School of Mathematics and Statistics, Lanzhou University, Lanzhou, China. [6]Oral and Maxillofacial Surgery, Faculty of Dentistry, The University of Hong Kong, Hong Kong, SAR, China. [7]School of Biomedical Engineering, Shanghai Jiao Tong University, Shanghai, China. [8]These authors contributed equally: Jiahao Bao, Huazhen Liu, Yu Zhuang, Leran Tao.[9]These authors jointly supervised this work: Shunyao Shen, Suncheng Xiang, Hongbo Yu. ✉e-mail: yhb3508@163.com

In clinical practice, some surgical planning software offers a soft tissue prediction function for visualizing treatment objective, which lacks sufficient accuracy to guide surgical skeletal movement decisions[9,10]. Various efforts have been made to achieve precise facial soft tissue deformation predictions in orthognathic surgery[11]. Current methodologies primarily include sparse landmark-based approaches and biomechanical modeling techniques. Sparse landmark-based approaches typically estimate bone-to-soft-tissue displacement ratios derived from clinical observations or machine learning algorithms, and they are commonly employed for predictions based on two-dimensional imaging data[12]. In contrast, biomechanical modeling approaches, including mass spring models, mass tensor models, and finite element models (FEM), facilitate simulations of complex and nonlinear tissue deformation patterns[13,14]. Among these, FEM effectively integrates the biomechanical properties of craniofacial tissues with patient-specific anatomical data. By utilizing precise interpolation methods to calculate the deformation gradient within each element, FEM robustly simulates complex nonlinear biomechanical behaviors, which is widely recognized as the most accurate and reliable technique for predicting facial soft-tissue changes[15,16]. Kim et al. developed a multi-stage FEM simulation method that incorporates realistic tissue sliding to enhance prediction accuracy[17]. They applied an incremental simulation approach with a realistic lip-sliding mechanism, significantly improving the prediction accuracy in the lip region[18]. Ruggiero et al. established a facial soft-tissue model incorporating detailed musculature structures and employed FEM to simulate postoperative facial morphology[19,20]. Although FEM demonstrates high prediction accuracy, a single simulation typically requires ~30 min to complete. Moreover, the complexity of FEM increases with anatomical detail, resulting in substantial computational demands and extended processing times, posing substantial challenges for meeting the demands of clinical applications[13,21].

Recently, the application of artificial intelligence (AI) has significantly transformed the clinical landscape of OGS. Many data-driven deep learning (DL) algorithms have been developed as a promising alternative for traditional biomechanical methods, achieving faster predictions with comparable accuracy[22–24]. Ter Horst et al. developed an autoencoder-based algorithm to predict soft tissue changes following mandibular advancement surgery. Ma et al. first proposed the FSC-Net to learn the nonlinear mapping from skeletal changes to facial shape responses[25]. Fang et al. proposed the Attentive Correspondence assisted Movement Transformation network (ACMT-Net) to predict postoperative facial shapes by calculating point-to-point attention correspondence matrices between bone and facial point sets, clarifying spatial relationships between facial soft tissue and bone[26,27]. Considering biomechanical properties, Lampen et al. developed a biomechanics-informed model based on the PointNet++ architecture, which integrates facial mesh geometry, bone segment displacement, and FEM boundary conditions as inputs to predict deformation. Nevertheless, current deep learning methods for facial prediction continue to face several notable limitations. First, previous studies frequently utilize small size datasets (~50 cases), overlooking the diverse range of deformity characteristics and surgical approaches, which limits the models' potential for generalization. Second, model interpretability remains insufficient, as the absence of biomechanical priors in geometric reasoning hampers the clarity of underlying predictive mechanisms for clinical decision-making. Finally, point cloud-based processing often leads to increased noise and information loss on the surface, compromising the quality of three-dimensional facial reconstructions.

To address the limitations identified in previous models, we developed a novel deep-learning framework named the Physics-Informed Skeletal-Facial Interaction Network (PhysSFI-Net) for orthognathic postoperative facial shape prediction, which was inspired by the biomechanical processes underlying soft tissue deformation. Through comprehensive validation using multiple quantitative metrics, our model has demonstrated superior performance compared to existing state-of-the-art methods. Our study contributes significantly in several aspects: (1) We introduced a hierarchical graph representation method to encode and extract geometric topological relationships between facial and skeletal structures and surgical plans, combined with an attention mechanism to predict the geometric features of facial displacement fields effectively. (2) We designed a novel Long Short-term Memory Networks (LSTM)-based soft tissue deformation prediction module, which accurately models the complex and continuous deformation processes induced by mechanical forces, thereby improving prediction accuracy. (3) Recognizing information loss inherent in point cloud down-sampling, we developed an electromechanics-informed high-precision reconstruction approach capable of rapidly generating postoperative facial meshes without compromising reconstruction accuracy.

## Results

### Study design

The overview of the study pipeline is shown in Fig. 1. In this study, we developed PhysSFI-Net, a physics-informed geometric deep learning framework for predicting postoperative facial morphology following orthognathic surgery. The model was trained and internally validated on a cohort of 135 patients and further evaluated using an independent multi-center dataset of 33 patients. Performance was assessed through objective quantitative metrics including global and subregional shape errors, mesh-based surface deviation, landmark localization error and subjective clinician-based evaluations. Comparative analyses with the state-of-the-art and baseline models, along with an ablation study examining the effect of LSTM depth, were conducted to comprehensively validate the effectiveness of the proposed approach.

### Participants and clinical characteristics

We retrospectively enrolled patients diagnosed with skeletal malocclusion from Department of Oral and Craniomaxillofacial Surgery, Shanghai Ninth People's Hospital, all of whom underwent comprehensive orthodontic and orthognathic combined treatment including treatment plan discussion, preoperative preparation, virtual surgical planning, and complete postoperative follow-up. Following the predefined inclusion and exclusion criteria, an internal dataset comprising 270 CT scans from 135 patients (preoperative-postoperative pairs) was established for model development and internal evaluation. Comprehensive clinical characteristics (gender, age, body mass index, skeletal discrepancy, facial asymmetry, Frankfort horizontal plane-mandibular plane angle) and surgery types (Lefort I osteotomy, bilateral sagittal split ramus osteotomy, genioplasty, paranasal bone grafting) are summarized in Fig. 2 and detailed in Supplementary Table S1. Among the participants, 68.1% were female, and the majority presented with a normal BMI ($18.5 \leq BMI \leq 25$). Consistent with the prevalence patterns observed in the Chinese population, Class III skeletal malocclusion accounted for the largest proportion of cases (73.3%), followed by Class II (17.8%). Facial asymmetry was identified in 61.5% of patients, and 56.3% exhibited a high-angle facial morphology (Fig. 2a). In terms of surgical types, 74 patients underwent a combined Le Fort I osteotomy and BSSRO, with 14 of these also receiving concomitant genioplasty (Fig. 2b, c). To improve midfacial projection, a subset of patients underwent paranasal bone grafting during orthognathic surgery. A schematic overview of the surgical procedures is provided in Fig. 2d.

### Experiments and model performance

The architecture of PhysSFI-Net was detailed in Materials and Methods section. The internal cohort ($n = 135$) was utilized to assess the model's predictive accuracy and to benchmark its performance against baseline (DGCNN and Point Transformer) and SOTA (ACMT-Net) methods[26–29]. Multiple quantitative metrics were employed for performance evaluation.

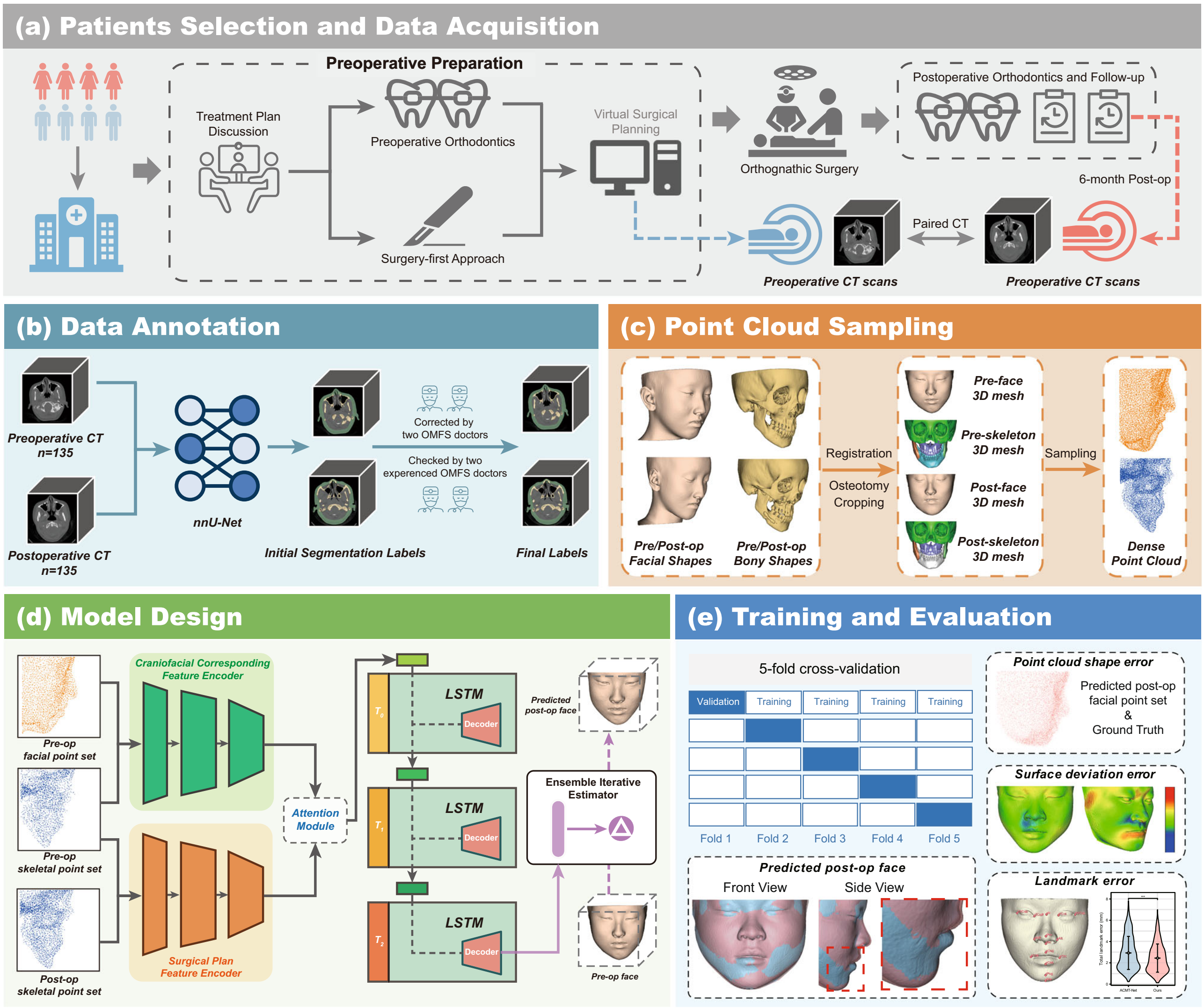

**Fig. 1 | The overview of the study pipeline.** **a** Patient selection and data acquisition. **b** Data annotation. **c** Point cloud sampling. **d** Model design. **e** Training and evaluation.

We first integrated the preoperative facial point cloud with both preoperative and postoperative skeletal point clouds in order to predict postoperative facial appearance. This enabled the generation of five sets of sparse predictions, which were subsequently fused into a unified dense point cloud representation. The Hausdorff Distance (HD) between the predicted postoperative facial point clouds and the corresponding ground-truth surfaces was calculated to quantify global shape discrepancies of 3D point clouds. Compared to the current SOTA model and baseline approaches, PhysSFI-Net consistently demonstrated lower prediction errors across all five sparse prediction sets (Fig. 3a; Supplementary Fig. S1). For the dense point cloud, PhysSFI-Net achieved the lowest HD (1.070 ± 0.088 mm), significantly outperforming ACMT-Net (1.186 ± 0.080 mm) and the baseline models including DGCNN (1.606 ± 0.287 mm) and Point Transformer (1.633 ± 0.280 mm) (Fig. 3b; Supplementary Table S2). Statistical analysis confirmed that the improvement of PhysSFI-Net over ACMT-Net was significant ($P < 0.05$) and the case-wise comparison further demonstrated the superior capability of the proposed method in capturing fine-grained facial morphology (Supplementary Fig. S2).

To further investigate the localized prediction accuracy, the dense facial point clouds were partitioned into five anatomically significant regions (chin, lip, nose, left cheek, and right cheek) and the corresponding subregional HD were calculated (Fig. 3c). As summarized in Supplementary Table S3, PhysSFI-Net consistently achieved the lowest prediction errors across all facial regions when compared with the comparative models. Specifically, the proposed method yielded mean HD values of 1.143 ± 0.086 mm for the chin, 1.149 ± 0.105 mm for the lip, 1.191 ± 0.089 mm for the nose, 1.201 ± 0.087 mm for the left cheek, and 1.194 ± 0.091 mm for the right cheek. In contrast, the SOTA model ACMT-Net exhibited higher errors across all regions, with particularly noticeable differences in the chin and cheek areas. The baseline models DGCNN and Point Transformer, demonstrated comparatively larger prediction errors, indicating their limited capability in capturing fine-grained facial deformations (Fig. 3d).

After completing the evaluation of point cloud shape accuracy, we further assessed the geometric fidelity of the reconstructed 3D facial surfaces. The predicted postoperative facial point clouds were converted into high-resolution surface meshes using the reconstruction framework described in the Materials and Methods section, enabling a more clinically meaningful evaluation of the predicted facial morphology. Figure 4 presents a qualitative comparison between the reconstructed 3D facial meshes generated by the comparative models and PhysSFI-Net and the corresponding ground-truth postoperative surfaces. Representative cases are shown in both frontal and lateral views, with magnified regions highlighting local morphological differences. Visually, PhysSFI-Net demonstrates closer agreement with the ground truth, particularly in anatomically critical regions such as the lips, chin, and nasal profile.

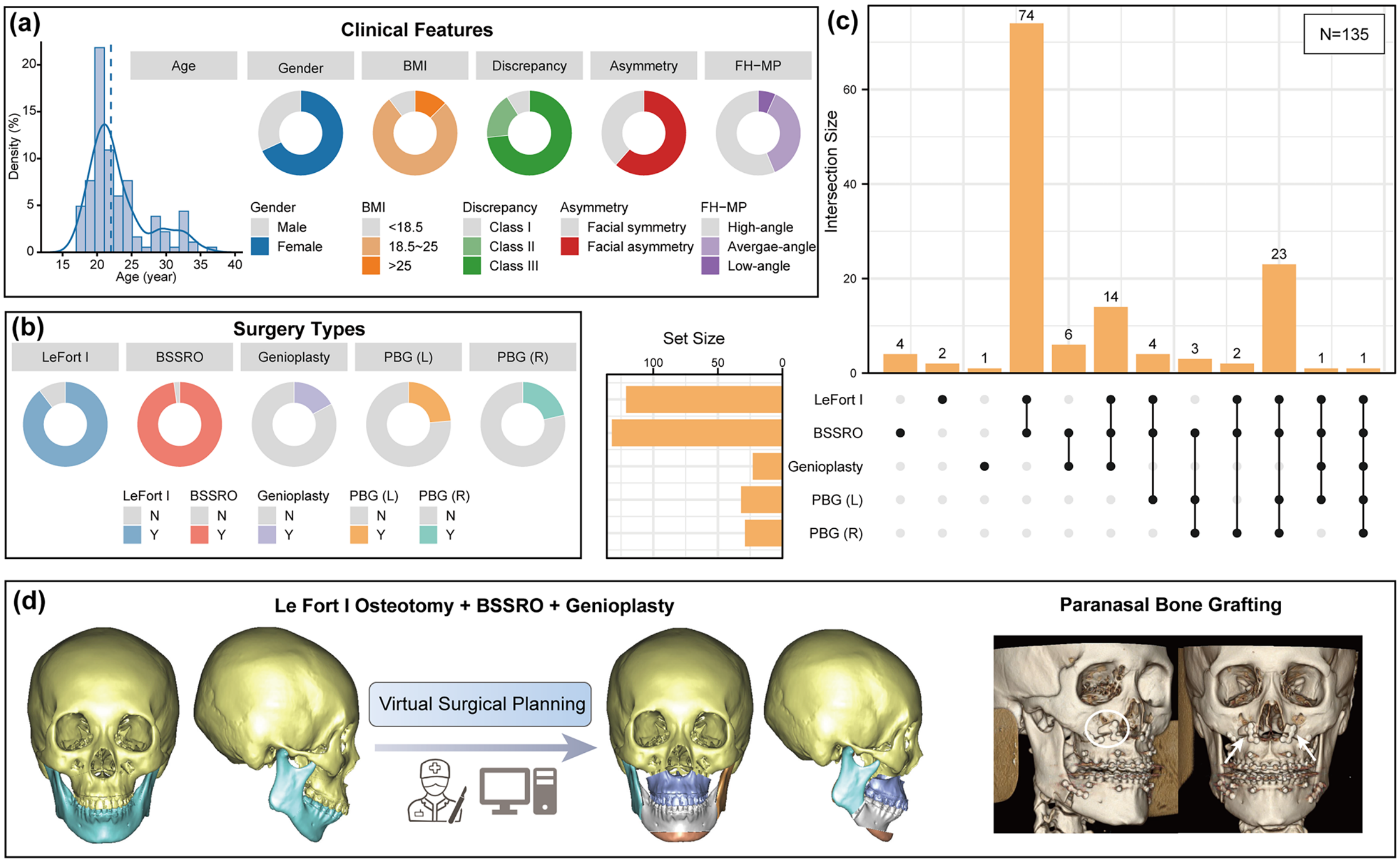

**Fig. 2 | Summary of patient characteristics and surgical procedures in the internal dataset.** **a** Clinical features of 135 patients, including age, gender, BMI, skeletal discrepancy, facial asymmetry, and vertical facial type. **b** Types of orthognathic procedures performed, including Le Fort I osteotomy, BSSRO, genioplasty, and paranasal bone grafting. **c** Distribution of surgical combinations across the dataset shown using an Upset plot. **d** Example of virtual surgical planning workflow.

To further illustrate the spatial distribution of surface deviation errors, heatmaps were generated by computing the signed surface deviation between the reconstructed predicted meshes and the ground-truth post-operative meshes (Fig. 5). These color-coded maps reveal that PhysSFI-Net exhibits more localized and lower-magnitude deviations compared with the state-of-the-art model ACMT-Net. Quantitative analysis demonstrated that PhysSFI-Net achieved a significantly lower average surface deviation than ACMT-Net (1.296 ± 0.349 mm vs. 1.372 ± 0.351 mm; $P < 0.05$) (Fig. 6a, b), further confirming the enhanced accuracy of the proposed method.

To assess anatomical fidelity, we evaluated the prediction accuracy across 15 clinically significant craniofacial landmarks, including five peri-orbital, four midfacial, and six perioral/chin landmarks (Fig. 7a; Supplementary Table S4). As shown in Fig. 7b, the overall landmark error for PhysSFI-Net was 2.445 ± 1.326 mm, significantly lower than that of ACMT-Net (2.930 ± 1.555 mm). At the individual landmark level, PhysSFI-Net consistently outperformed ACMT-Net, with statistically significant improvements observed at key landmarks including sICaL, sOCaR, sPrn, sSn, sAIL, sAIR, sChU, sChL, sChR, sPog, and sMe (Fig. 7c). Given that a landmark error of less than 2 mm is generally regarded as clinically acceptable, we further examined the distribution of landmark errors using 2 mm and 4 mm as thresholds. PhysSFI-Net yielded a significantly higher proportion of landmarks with errors below 2 mm and a lower proportion exceeding 4 mm when compared to ACMT-Net (Fig. 7d), demonstrating its superior clinical relevance and robustness.

## Multi-center validation

As illustrated in Fig. 8a, the multicenter external validation was performed. The cohort comprised 33 patients from three independent institutions and included a diverse range of dentofacial deformities and surgical procedures, predominantly skeletal Class III malocclusion treated with Le Fort I osteotomy and BSSRO (Fig. 8b and Supplementary Table S5). The mean height and weight were 167.97 ± 9.70 cm and 62.56 ± 14.29 kg, with a median age of 23 years (Fig. 8c). The distribution of patient-reported chief complaints, primarily related to facial aesthetics and occlusal dysfunction, is presented in Fig. 8d.

Objective evaluation in the multicenter cohort demonstrated the superior performance of PhysSFI-Net. The global HD was significantly reduced compared with ACMT-Net (1.431 ± 0.087 mm vs. 1.988 ± 0.296 mm). Subregional analysis further revealed consistently lower errors for PhysSFI-Net across all anatomically significant facial regions, including the chin (1.400 ± 0.112 mm), lip (1.432 ± 0.108 mm), nose (1.444 ± 0.098 mm), left cheek (1.443 ± 0.095 mm), and right cheek (1.461 ± 0.083 mm), with the most pronounced improvements observed in the chin and cheek regions. Furthermore, mesh-based evaluation revealed a significantly lower average surface deviation for PhysSFI-Net (1.774 ± 0.121 mm) than for ACMT-Net (2.009 ± 0.161 mm; $P < 0.05$), confirming its superior geometric fidelity and clinical applicability.

For subjective evaluations, representative qualitative comparisons showed that the predicted postoperative facial morphology closely resembled the ground truth in both frontal and lateral views (Fig. 8h). In the clinical acceptability assessment, PhysSFI-Net achieved a higher proportion of clinically acceptable predictions and markedly fewer inaccurate cases in both frontal and lateral views (Fig. 8i). In the frontal view, 19 cases were rated as clinically acceptable and only 2 as inaccurate, compared with 8 and 11 cases for ACMT-Net. Similar improvements were observed in the lateral view. Consistently, the reasoning consistency analysis (Fig. 8j) revealed greater agreement with surgeons' expectations for PhysSFI-Net, with more highly consistent and fewer inconsistent evaluations in both frontal (20 vs. 9 highly consistent; 4 vs. 12 inconsistent) and lateral views (16 vs. 10 highly consistent; 6 vs. 9 inconsistent).

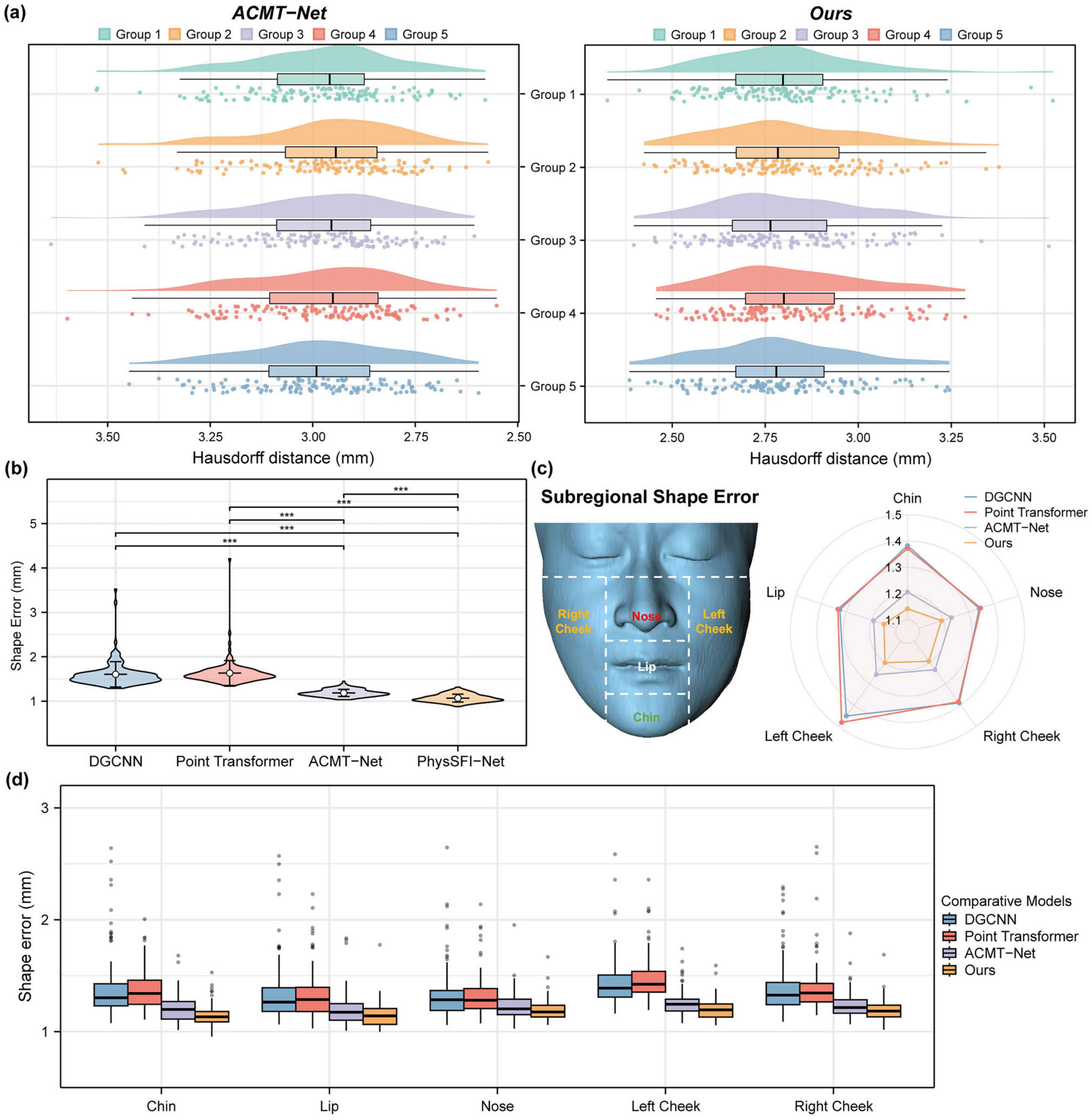

**Fig. 3 | Comparison of global and subregional point cloud shape errors among the comparative models and PhysSFI-Net.** **a** Hausdorff distance (HD) distributions for five groups of sparse point cloud subsets, comparing ACMT-Net (left) and PhysSFI-Net (right). **b** Global shape error on dense point clouds, showing a significant reduction in HD with PhysSFI-Net. **c** Schematic illustration of facial subregional partitioning and radar plot comparing regional shape errors (chin, lips, nose, left cheek, and right cheek). **d** Boxplot comparison of subregional shape errors across all models.

## Effect of LSTM-based sequential prediction

An ablation study was performed to evaluate the effect of LSTM-based sequential prediction by varying the number of LSTM layers from 0 to 4. As shown in Supplementary Fig. S3 and Supplementary Table S7, the results demonstrated a progressive reduction in HD from 1.129 ± 0.088 mm (LSTM-0) to an optimal value of 1.071 ± 0.068 mm with three LSTM layers (LSTM-3). No significant difference was observed between LSTM-2 and LSTM-3, while the performance deteriorated when increasing to four layers (1.170 ± 0.098 mm). Training and testing loss curves indicated stable convergence across all configurations, with LSTM-2 and LSTM-3 exhibiting superior generalization (Supplementary Fig. S4). These findings confirm the effectiveness of sequential modeling and support the selection of three LSTM layers in the final model.

Grad-CAM visualization was employed to interpret the prediction mechanism of PhysSFI-Net (Supplementary Fig. S5). The resulting attention maps highlighted anatomically significant regions of the preoperative facial point cloud, particularly the lips and chin, indicating that the model captures clinically meaningful skeletal–facial interactions. This qualitative analysis supports the interpretability and clinical relevance of the proposed framework.

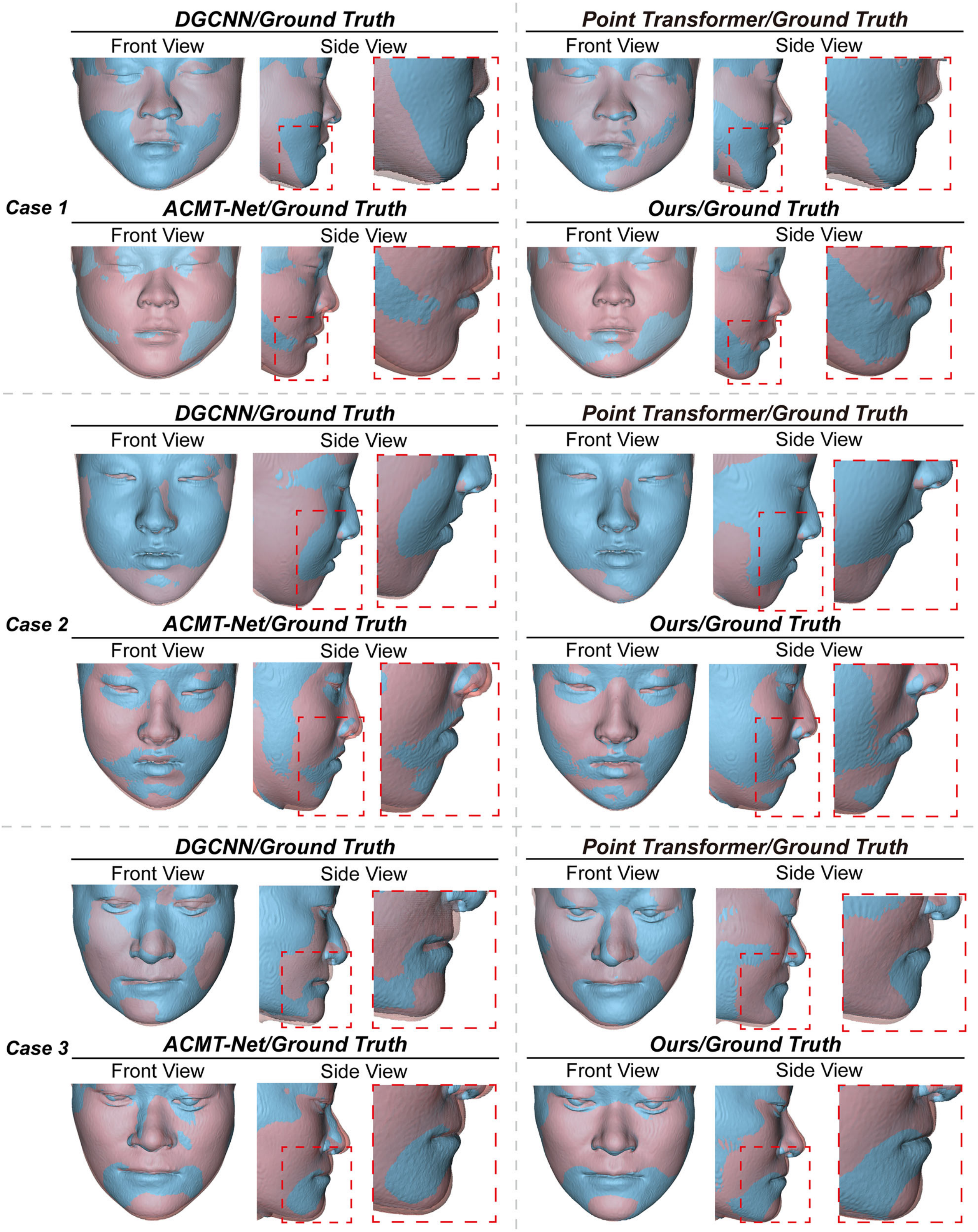

**Fig. 4 | Qualitative comparison of postoperative facial appearance predictions among comparative models and the proposed PhysSFI-Net.** Three representative cases are shown with front and side views. Red surfaces represent the ground truth, and blue surfaces represent the predicted results.

## Discussion

Prediction of postoperative facial appearance is critical for optimizing orthognathic surgical plans[30,31]. Conventional approaches integrated into existing clinical software often suffer from limited predictive precision. The development of fast and accurate prediction methods remains an unmet need. In this study, we propose a physics-informed geometric deep learning framework, named PhysSFI-Net. We designed a hierarchical graph representation method to encode the geometric and topological relationships among facial soft tissues, skeletal anatomy, and planned surgical movements. It incorporates an attention mechanism to estimate the postoperative facial displacement field and integrates an LSTM-based soft-tissue deformation module to capture the temporally continuous and biomechanically driven nature of soft tissue response. Comprehensive evaluations using multiple quantitative metrics confirm its superior performance relative to existing SOTA approaches.

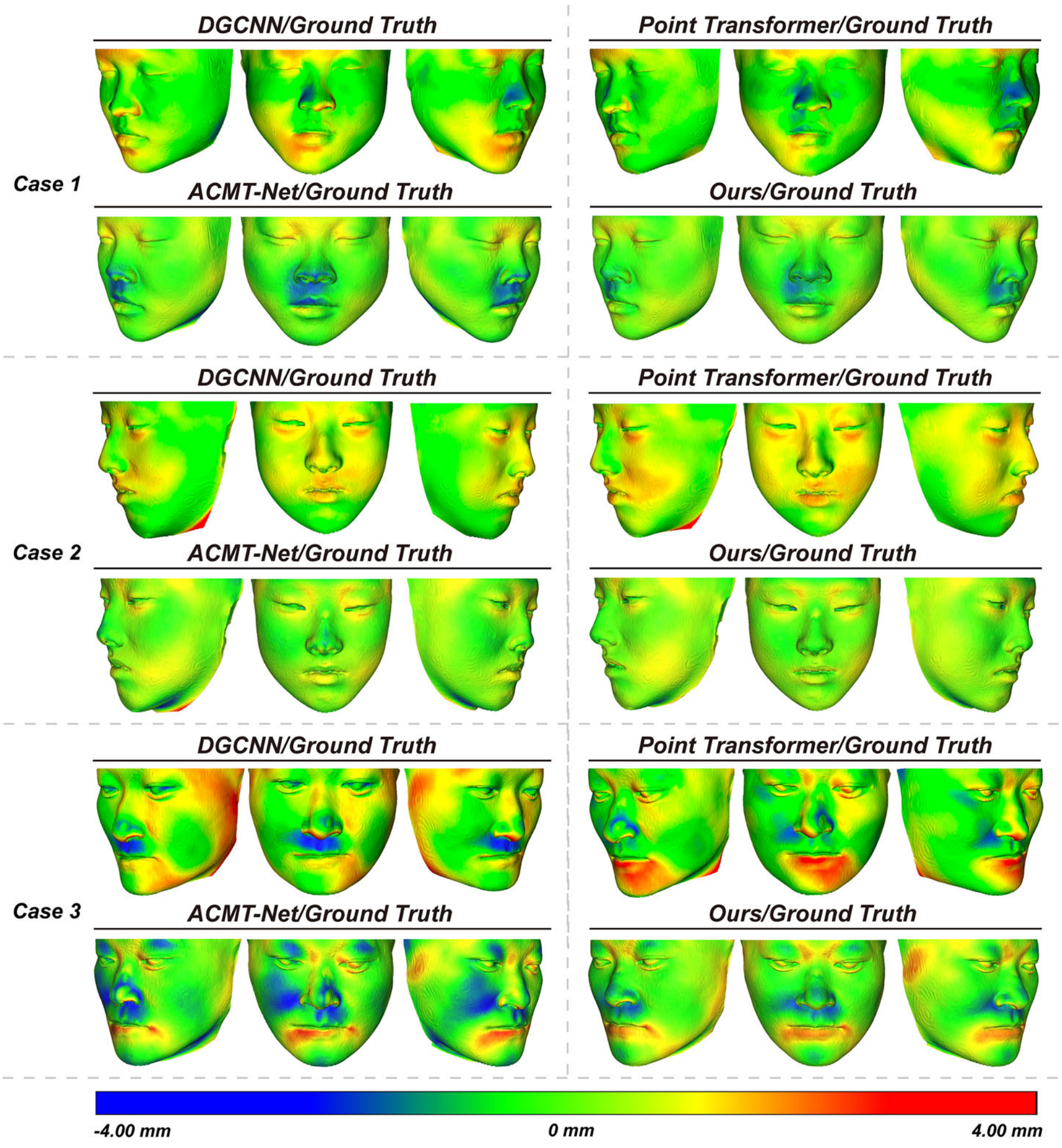

**Fig. 5 | Heatmap visualizations of surface deviation errors for representative cases.** The color scale indicates the magnitude of the signed surface deviation, ranging from −4.00 mm (blue) to +4.00 mm (red), with green representing minimal deviation.

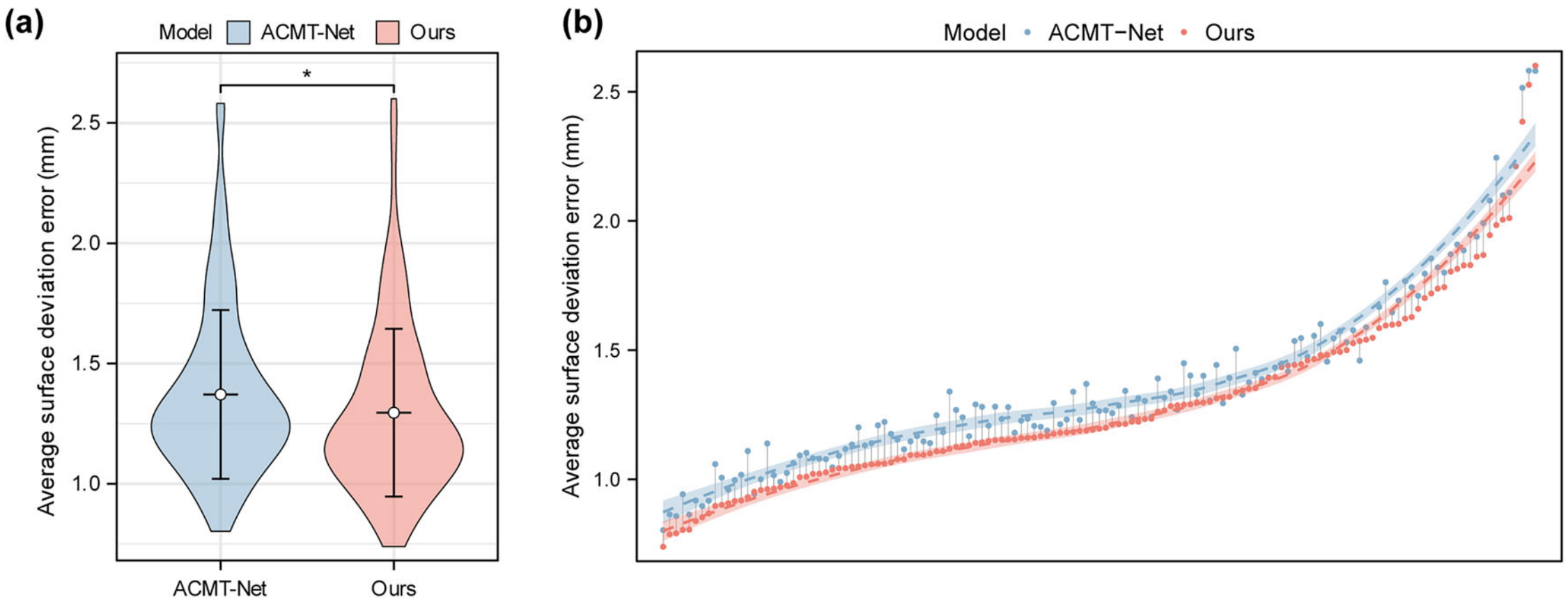

**Fig. 6 | Quantitative evaluation of surface deviation errors between ACMT-Net and the proposed model. a** Violin plots of average surface deviation error show that our model achieves significantly lower errors than ACMT-Net (*$p < 0.05$). **b** Case-wise paired comparison of average surface deviation error, with most samples showing reduced error using our method.

Biomechanical simulations (e.g., finite element modeling) can produce realistic results but often struggle to balance accuracy and speed, making routine clinical use difficult[17,18]. To overcome these limitations, researchers have turned to geometry-based deep learning frameworks that learn the mapping from skeletal movements to soft-tissue deformation directly from patient data. Notable examples include FSC-Net, DGCFP and ACMT-Net, which leverage neural networks to extract bone displacement features and infer soft-tissue deformations[25,26,32]. For instance, Huang et al. proposed DGCFP for postoperative facial prediction, which consists of multi-scale dualconv face encoder, pointwise bone encoder, dual-space movement transfer and coarse-to-fine deformation[32]. ACMT-Net introduced an attentive bone–soft-tissue correspondence mechanism to achieve FEM-level accuracy with significantly improved efficiency[26]. Based on these advances, our proposed PhysSFI-Net provides both rapid inference and high predictive accuracy, enabling postoperative facial predictions within seconds and supporting practical implementation in routine clinical settings. The model achieved a dense point cloud shape error of 1.070 ± 0.088 mm and a mesh reconstruction error of 1.296 ± 0.349 mm. Beyond global metrics, PhysSFI-Net demonstrated consistent improvements in anatomically critical regions, including the chin, lip, nose, left cheek and right cheek. Comparative experiments further showed that PhysSFI-Net outperformed both the SOTA model ACMT-Net and baseline models DGCNN and Point Transformer, particularly in capturing fine-grained anatomical details in aesthetically sensitive areas. These advantages can be attributed to the integration of hierarchical graph representations and an LSTM-based biomechanics-inspired sequential deformation module, enabling more accurate modeling of the dynamic interaction between skeletal repositioning and soft-tissue adaptation.

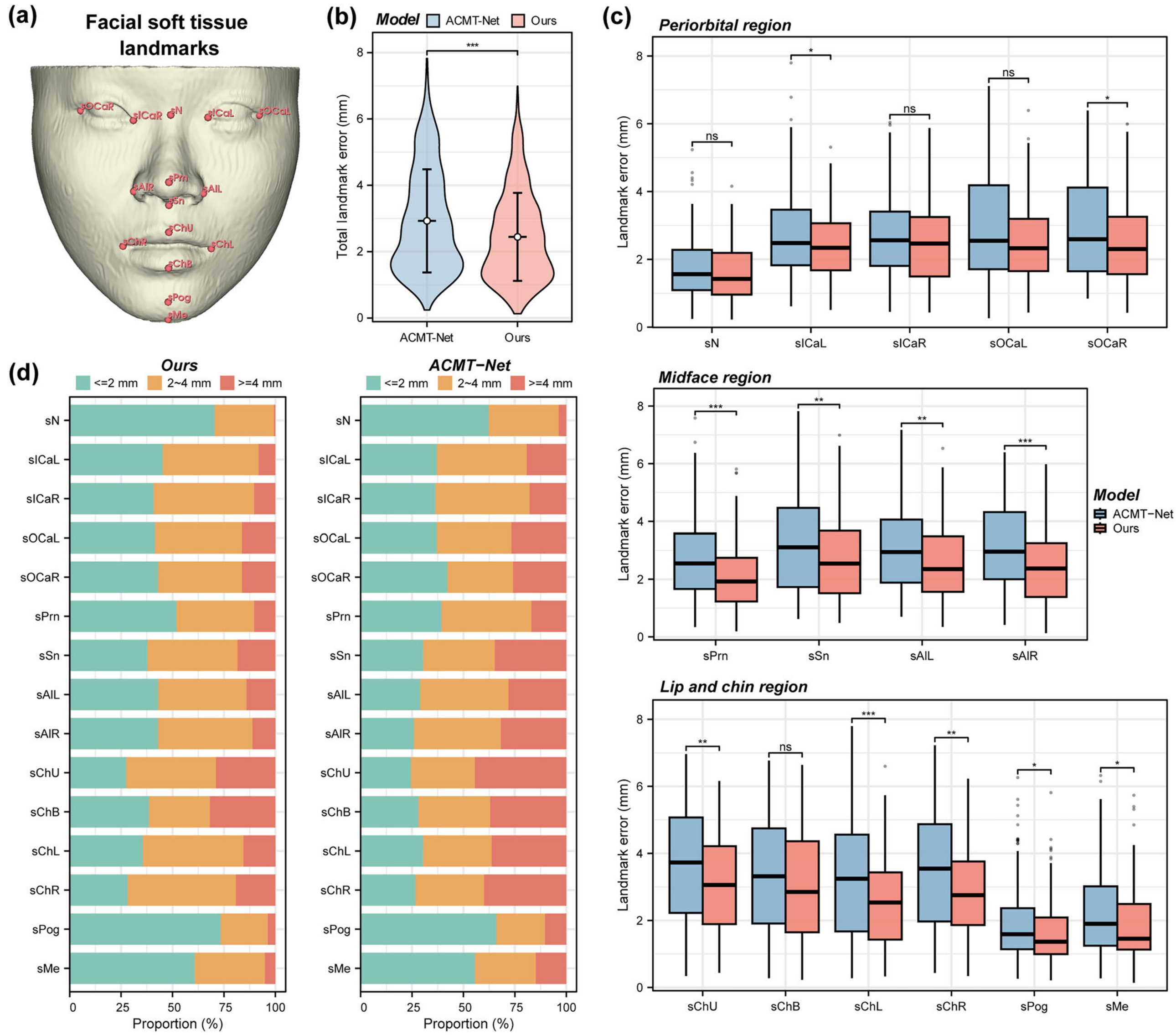

**Fig. 7 | Landmark-based evaluation of soft-tissue prediction accuracy.** **a** Definition of 15 facial soft-tissue landmarks used for accuracy evaluation. **b** Violin plots showing total landmark error across all points, with our model demonstrating significantly lower errors than ACMT-Net (***$p < 0.001$). **c** Landmark error comparison across periorbital, midface, and lip/chin regions. **d** Proportion of predicted landmarks within three error ranges (<2 mm, 2–4 mm, and >4 mm), illustrating the overall improvement in accuracy with our model.

The motivation for incorporating an LSTM module in PhysSFI-Net arises from the recognition that facial soft-tissue deformation is fundamentally biomechanical and involves incremental changes under mechanical forces[21]. Purely geometric approaches typically model static shape correspondences, potentially overlooking the dynamic and cumulative nature of soft-tissue deformation driven by biomechanical interactions. From a clinical perspective, patient-specific biomechanical parameters such as tissue elasticity and muscle configuration vary considerably but are difficult to obtain in routine practice[33]. Moreover, constructing explicit biomechanical models is time-consuming and computationally demanding, limiting their clinical applicability. Therefore, rather than developing a fully physics-based model, our approach integrates biomechanics-inspired principles into a data-driven framework. PhysSFI-Net leverages geometric information readily available in clinical workflows while emphasizing the mechanics-driven interaction between skeletal repositioning and soft-tissue deformation. Previous studies have attempted to integrate physical principles into deep learning frameworks to improve model interpretability by incorporating physics-informed constraints. Lampen et al. introduced a biomechanics-informed deep neural network based on the PointNet++ and subsequently proposed a DL method named Spatiotemporal Incremental Mechanics Modeling based on PhysGNN, to perform spatiotemporal incremental simulations of soft tissue mechanical modeling[14,21,30,34]. LSTM is a type of recurrent neural network designed to effectively model sequential data and capture long-range dependencies[35]. Karami et al. combined CNN and LSTM layers with a mass-conservation loss to simulate viscoelastic tissue behavior[36]. Likewise, Nguyen-Le et al. employed an LSTM-based network to predict pelvic soft-tissue deformation in childbirth simulation[37]. These approaches show that LSTM can capture nonlinear, time-dependent mechanical behavior by learning how deformations evolve in sequence. Building on this concept, PhysSFI-Net incorporates an LSTM-based incremental deformation module, enabling the network to simulate soft-tissue deformation through a sequence of cumulative steps driven by mechanical forces. This stepwise approach significantly improves the model's interpretability, as each incremental LSTM step represents an intermediate stage of deformation, thereby ensuring biomechanical plausibility and reducing physically unrealistic distortions.

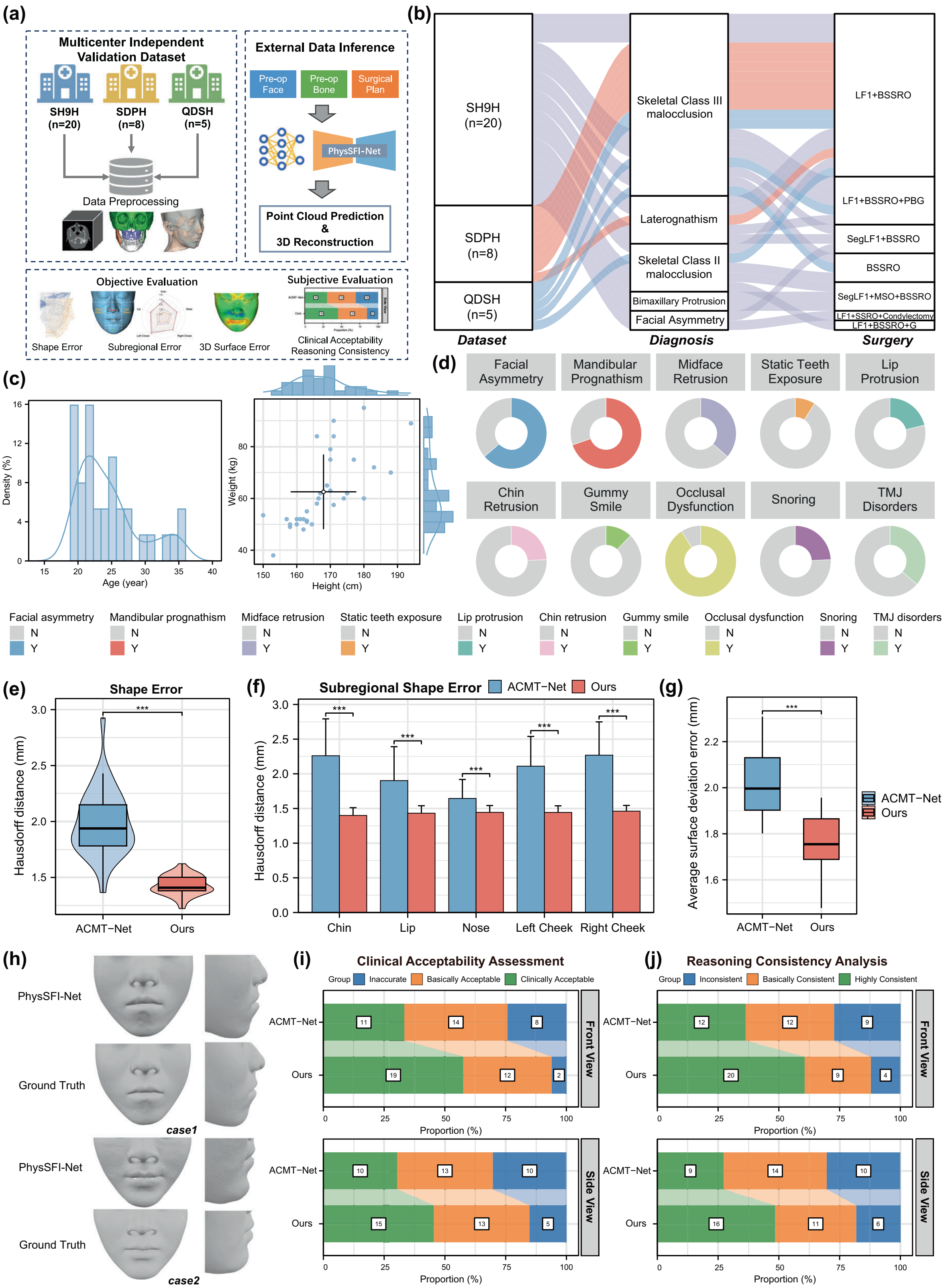

**Fig. 8 | Multicenter external validation of PhysSFI-Net. a** Overview of the external validation workflow, including data collection from three institutions, preprocessing, model inference, and objective and subjective evaluations. **b** Distribution of datasets, diagnostic categories, and surgical procedures across the multicenter cohort. **c** Demographic characteristics of the external cohort, including age, weight and height distributions. **d** Distribution of chief complaints among patients. **e** Comparison of global point cloud shape error (Hausdorff distance) between ACMT-Net and PhysSFI-Net. **f** Subregional shape errors across five facial regions (chin, lip, nose, left cheek, and right cheek). **g** Comparison of average surface deviation between reconstructed meshes. **h** Representative qualitative comparisons of predicted postoperative facial morphology and ground truth. **i** Clinical acceptability assessment by experienced surgeons. **j** Reasoning consistency analysis evaluating the agreement between predicted outcomes and surgeons' expectations.

Within the broader landscape of prediction approaches in orthognathic surgery planning, 3D soft-tissue outcome modeling is widely acknowledged as one of the most difficult aspects. Numerous recent approaches have been developed to address this challenge from diverse methodological perspectives. Kim et al. introduced a graph learning-based prediction method using lateral cephalograms, named Dual Embedding Module Graph Convolutional Neural Network, which was developed to predict the displacement of key skeletal landmarks such as ANS, PNS, B-point and Md1crown[38]. Building on that, they recently developed GPOSC-Net, a generative approach that first predicts postoperative landmark shifts with a graph neural network and then synthesizes a realistic postoperative lateral cephalogram using a latent diffusion model[39]. Other researchers have focused on parametric modeling of the face. Qiu et al. developed SCULPTOR, a skeleton-consistent face generator that jointly models the skull and facial surface in a unified data-driven framework, which can produce anatomically plausible facial modifications and simulate surgical outcomes[40]. In addition, Han et al. proposed an automated pipeline based on the FLAME 3D morphable model to predict postoperative facial appearance, reporting mean errors of ~9 mm in Hausdorff distance and 2.5 mm in Chamfer distance[41]. Compared to these methods, PhysSFI-Net directly predicts the complete 3D soft-tissue morphology rather than being limited to cephalometric projections, resulting in predictions closer to actual postoperative outcomes. Its physics-informed LSTM architecture enhances interpretability and biomechanical realism, addressing limitations inherent in purely statistical or generative approaches. These features collectively position PhysSFI-Net as a comprehensive and clinically relevant approach for orthognathic surgical outcome simulation.

The other major strengths of this study lie in the sample size of the dataset used for model training and validation. Previous studies often relied on small 3D datasets, typically comprising ~40 paired samples, which restrict the generalizability and stability of predictive models. Some studies have utilized synthetic data generated from finite element simulations. While such data incorporate biomechanical assumptions, they do not reflect the anatomical variability present in real clinical cases. To address these limitations, we constructed one of the largest real-world datasets reported to date, consisting of 135 orthognathic surgery patients with paired preoperative and postoperative CT scans for model training and internal validation. This dataset includes a wide spectrum of skeletal deformities and surgical plans, providing strong representativeness, diversity, and anatomical completeness. Furthermore, we applied a five-fold cross-validation approach to rigorously assess model performance across different subsets of the data. This strategy enhances the robustness and generalizability of the proposed PhysSFI-Net framework and supports its potential for clinical application. Importantly, we additionally incorporated an independent multicenter external validation cohort comprising 33 patients from three medical institutions. This heterogeneous cohort enabled an unbiased evaluation of model performance across different clinical environments and imaging protocols, substantially strengthening the external validity and clinical applicability of PhysSFI-Net. Collectively, the combination of a large-scale internal dataset and independent multicenter validation provides a solid foundation for the reliability, generalizability, and translational potential of the proposed framework[42].

Our study has several limitations. Although the model was trained on a relatively large single-center dataset, a preliminary multicenter external validation involving patients from three independent institutions demonstrated consistent performance across diverse clinical settings. Future studies incorporating larger-scale, multicenter cohorts with patients from diverse ethnic backgrounds and a broader spectrum of dentofacial deformities are warranted to further enhance the robustness and global applicability of the proposed framework. In addition, soft-tissue reconstructions generated from CT data lack critical color and texture information. Second, our current prediction framework does not adequately incorporate patient-specific characteristics, and several preprocessing steps still rely on manual intervention. Achieving a fully automated, end-to-end predictive pipeline remains a significant objective to enhance clinical applicability. Third, accurately predicting soft-tissue deformation in anatomically complex regions, particularly around the lips, remains a significant challenge. Factors such as surgical suture techniques, intraoral orthodontic brackets and patient movements during CT imaging can substantially influence prediction accuracy. Developing standardized methods to represent individual soft-tissue characteristics is essential for improving model robustness and precision. Furthermore, incorporating more explicit biomechanical response mechanisms in future studies may enhance both the interpretability and physiological plausibility of the proposed framework.

In conclusion, we proposed PhysSFI-Net, a physics-informed deep learning framework that integrates hierarchical graph representations, attention-based feature encoding, and an LSTM-driven soft-tissue deformation module to simulate biomechanically realistic facial changes. Our model demonstrated high prediction accuracy across multiple quantitative metrics and outperformed the current state-of-the-art method, highlighting its potential value in advancing personalized orthognathic surgical planning and decision support in clinical practice.

## Methods

Our study adheres to the Checklist for Artificial Intelligence in Dental Research. The study was performed in accordance with the Declaration of Helsinki and was approved by the ethics committee of Shanghai Ninth People's Hospital, Shanghai Jiao Tong University School of Medicine (IRB No. SH9H-2022-TK12-1). Informed consent was obtained from all participants.

### Data acquisition

This study retrospectively enrolled patients diagnosed with skeletal malocclusion from Department of Oral and Craniomaxillofacial Surgery, Shanghai Ninth People's Hospital, all of whom underwent comprehensive orthodontic and orthognathic combined treatment including treatment plan discussion, preoperative preparation, virtual surgical planning, and complete postoperative follow-up. Patients with congenital dentofacial deformities ($n = 12$), those who underwent additional soft-tissue cosmetic procedures ($n = 18$), or those with facial prostheses such as polyether ether ketone (PEEK) implants ($n = 15$) were excluded from the study. All CMF CT scans were required to meet predefined quality standards. Cases with severe metal artifacts or poor overall image quality were excluded ($n = 10$). Preoperative CT data were acquired during the virtual surgical planning phase, ~1 month before surgery, when patients had completed preoperative orthodontic treatment and their teeth and skeletal structures were stabilized. Paired postoperative CT scans were collected six months after surgery, at which point soft-tissue swelling had fully resolved (Fig. 9a). A total of 135

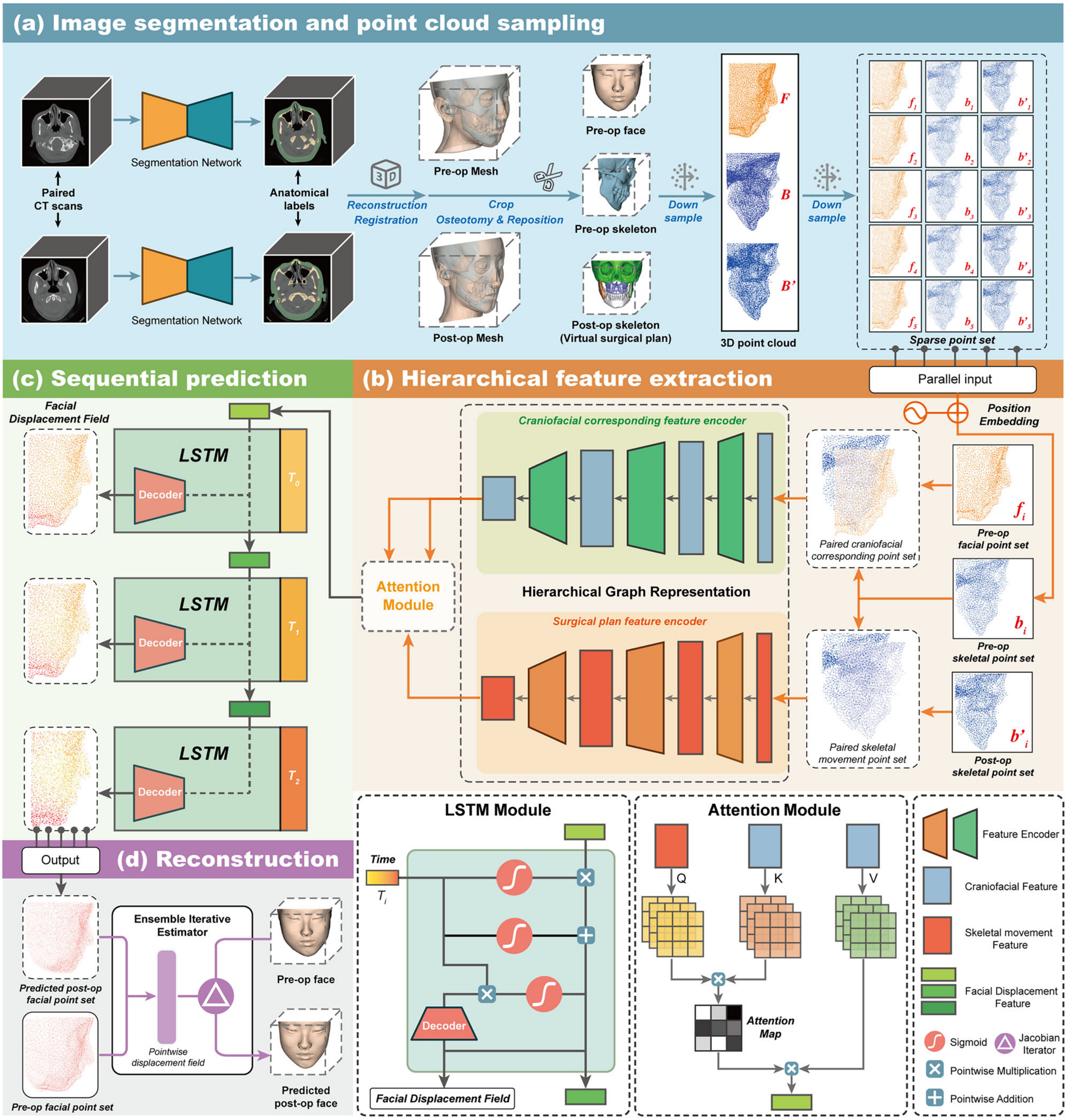

**Fig. 9 | The detailed architecture of Physics-Informed Skeletal-Facial Interaction Network (PhysSFI-Net).** **a** Image segmentation and point cloud sampling. **b** A hierarchical feature extraction approach, consisting of a craniofacial correspondence feature encoder and a surgical plan feature encoder, combined with an attention mechanism designed to predict facial displacement features. **c** A sequential prediction approach that employs three LSTM modules to simulate incremental soft tissue deformation guided by mechanical states. **d** An electromechanics-inspired high-precision reconstruction method, facilitating accurate 3D reconstruction of postoperative facial morphology.

patients who underwent orthognathic surgery were included for model training and validation.

To further evaluate the robustness and clinical feasibility of the proposed model, an independent multi-center dataset was constructed using the same inclusion and exclusion criteria as the primary cohort. This dataset included 20 patients from different clinical teams at Shanghai Ninth People's Hospital (SH9H), 8 patients from Shandong Provincial Hospital (SDPH), and 5 patients from Qingdao Stomatological Hospital (QDSH). All data were collected and preprocessed following the same protocol as the training dataset. Notably, this external cohort was not used for model training. Instead, it was solely employed for external inference and performance evaluation, enabling an unbiased assessment of the model's generalizability across different institutions.

## Data annotation and preprocessing

CT scans were obtained as DICOM format. A semi-automated workflow was utilized for segmenting craniofacial skeletal and soft-tissue structures. A pretrained nnU-Net model was employed to generate initial segmentation masks, which were then carefully refined slice by slice at the voxel level by two oral and maxillofacial surgeons. After verification by two expert clinicians, the final segmentation results were confirmed (Fig. 9b). Due to variations in patient positioning and the extent of CT coverage, each segmentation was cropped to retain only regions relevant to orthognathic procedures. All tissues posterior to the anterior margin of the external auditory canal and superior to the supraorbital rim were deleted using 3D Slicer software (version 5.0.2). Then, all patient skeleton and face data were registered to a unified coordinate system. For each patient, we employed an Iterative Closest Point (ICP) algorithm to align paired preoperative and postoperative model using stable landmarks (e.g., nasion, medial canthus, and lateral canthus). A paired dataset was established which consists of preoperative and postoperative craniofacial skeletal models and soft tissue surfaces.

To simulate a clinically relevant VSP scenario while preserving the true correspondence between postoperative skeletal structures and soft tissues, postoperative CT data were utilized as a bridge to reconstruct a planning-consistent postoperative skeletal configuration derived from the preoperative anatomy. We acquired pointwise displacement data of skeletal segments through the following methodological workflow: (1) The preoperative skeletal model was imported into Mimics software and segmented according to standard orthognathic surgical procedures. The segmentation resulted in distinct cranial, maxillary, mandibular body, and bilateral mandibular ramus segments. For patients who underwent genioplasty, an additional chin segment was also isolated. (2) The postoperative skeletal model was then imported and precisely registered to the preoperative model utilizing cranial anatomical landmarks as references. The preoperative skeletal segments were individually translated and rotated, ensuring each segment accurately aligned with the corresponding postoperative segment positions. (3) The repositioned skeletal segments were merged with the cranial segment to generate a surgical planning skeletal model. This approach ensured pointwise correspondence in the sampled bone point clouds and maintained complete conformity with the postoperative skeletal anatomy. To balance geometric fidelity and computational efficiency, both skeletal and facial soft-tissue surface meshes were represented using point clouds[22]. High-resolution 3D meshes were first randomly downsampled to 20,480 points to form dense point clouds. Subsequently, Farthest Point Sampling (FPS) was employed to generate five mutually complementary sparse point cloud subsets, enabling efficient multi-scale feature learning while preserving essential anatomical details[43,44].

### Overall architecture of PhysSFI-Net

The overall architecture of our proposed model, named PhysSFI-Net, is illustrated in Fig. 9b–d. PhysSFI-Net comprises three primary components: (1) A hierarchical graph representation module, consisting of a craniofacial correspondence feature encoder and a surgical plan feature encoder, combined with an attention mechanism designed to predict facial displacement features (Fig. 9b). (2) A sequential prediction approach that employs three Long Short-term Memory (LSTM) Networks modules to simulate incremental soft tissue deformation guided by mechanical states (Fig. 9c). (3) An electromechanics-inspired high-precision reconstruction method, facilitating accurate 3D reconstruction of postoperative facial morphology (Fig. 9d).

We formulate the postoperative facial shape prediction task using point clouds as the primary data structure. Given the preoperative skeletal shape $B = \{b_i \in \mathbb{R}^3\}_{i=1}^{N_B}$, the postoperative skeletal shape $B' = \{b'_i \in \mathbb{R}^3\}_{i=1}^{N_B}$, and the preoperative facial surface $F = \{f_j \in \mathbb{R}^3\}_{j=1}^{N_F}$, the goal is to predict the postoperative facial surface $F' = \{f'_j \in \mathbb{R}^3\}_{j=1}^{N_F}$ by learning a mapping $\Phi$:

$$F' = \Phi(B, B', F) = F + \Delta F(B, B', F) \tag{1}$$

To comprehensively and accurately capture the structured information embedded within the point cloud data, we adopt a geometric-topological perspective by modeling the point cloud as a 2D manifold embedded in 3D space. This modeling paradigm aims to explicitly characterize the local manifold properties of point clouds, thereby enabling the extraction of rich geometric structures inherently encoded in the data. Accordingly, the mapping $\Phi$ can be interpreted as a transformation between manifolds, facilitating a geometry-aware prediction of soft tissue deformation:

$$\Phi : \mathcal{M}_B \times \mathcal{M}_{B'} \times \mathcal{M}_F \rightarrow \mathcal{M}_{F'} \tag{2}$$

In this context, Eq. (1) can be regarded as the discrete sampling form of the underlying continuous manifold mapping.

### Hierarchical feature extraction

To enhance computational efficiency, the input point cloud is partitioned into five low-resolution sub-clouds, which are processed in parallel. After obtaining the predicted displacement fields for each sub-cloud, we perform high-resolution reconstruction to generate the final output. Each sub-cloud consists of three components: the preoperative skeletal point cloud, the postoperative skeletal point cloud, and the preoperative facial point cloud. These heterogeneous point clouds are uniformly represented as a label-augmented manifold structure, which we define as an enhanced manifold representation. This representation encodes both geometric information and semantic identity (e.g., anatomical source) of each point, enabling the network to jointly learn across structurally distinct yet spatially correlated modalities:

$$\mathcal{M}(p_i) = \left\{ (g_i, \ell_i) | p_i \in \mathbb{R}^3, \ell_i \in \{0,1\}^3 \right\} \tag{3}$$

where $P_i$ denotes the 3D coordinates of skeletal or facial point clouds, while $g_i$ is a function of the input location, encoding the local graph structure of the underlying manifold:

$$g_i = \sum\nolimits_{j \in \mathcal{N}(i)} w_{ij} \phi(p_j, p_i - p_j), \mathcal{N}(i) = KNN(p_i) \tag{4}$$

Here, the label $I_i$ serves as a positional encoding for each point in the point cloud, providing additional spatial context to guide the learning of manifold-aware features:

$$l_i(p_i) = \begin{cases} \sin\left(\frac{p_i}{10000^{2k/C}}\right) & i = 2k \quad , \\ \cos\left(\frac{p_i}{1000^{2k/C}}\right) & i = 2k+1 \quad . \end{cases} \tag{5}$$

where $k = 0, 1, 2, \ldots, \lfloor C/2 \rfloor - 1$. Heterogeneous point clouds are treated as an integrated entity rather than being processed separately, enabling more effective capture of local geometric relationships across different anatomical structures. The network consists of two primary components: feature extraction and facial surface reconstruction. The feature extraction module (Fig. 9b) is designed to encode skeletal displacement and model the skeletal-to-facial correspondence. A hierarchical PointNet-based architecture is employed to extract structural features, while graph convolutional layers are incorporated to further capture local topological dependencies (Supplementary Fig. S6)[45]. Upon obtaining the encoded skeletal displacement features and skeletal-facial relational features, we introduce a multi-head attention mechanism to perform hierarchical feature extraction. In each attention layer, the facial surface features serve as the query ($Q_h$), the skeletal-facial relational features as the key ($K_h$), and the skeletal displacement features as the value ($V_h$). Attention maps are computed to aggregate geometry-aware features, resulting in refined representations for facial shape prediction:

$$\boldsymbol{A}_h = \text{softmax}\left(\frac{\boldsymbol{Q}_h \boldsymbol{K}_h^T}{\sqrt{d_h}} + \boldsymbol{E}_h\right)\boldsymbol{V}_h \tag{6}$$

Here, $E_h$ denotes the relative positional encoding, which introduces geometric priors into the attention mechanism by capturing spatial relationships between query and key points.

### Sequential prediction based on LSTM

To simulate the physically constrained postoperative deformation process, we employ an LSTM-based decoder that iteratively generates the facial displacement field (Fig. 9c). At each time step $t \in \{0, 1, \ldots, T-1\}$, the decoder outputs an incremental displacement vector $\delta^t \in \mathbb{R}^{N_F \times 3}$. The final predicted displacement field is then obtained by summing the outputs across all time steps:

$$\Delta f = \sum_t \delta^t \tag{7}$$

The iterative process of the LSTM decoder is designed to simulate the continuous evolution of soft tissue deformation. At each time step, the decoder takes the current time index, feature representation, and intermediate displacement field as input, and outputs the updated feature state and displacement increment. Formally, the process can be described as:

$$\left[x^t; \delta^t\right] = LSTM\left[t; x^{t-1}; \delta^{t-1}\right] \tag{8}$$

The iterative process is initialized at t = 0, $x^0 = \bigoplus_h A_h$, and $\delta^0 = 0$.

### High-resolution reconstruction

High-resolution reconstruction of post-operative facial soft tissue requires balancing physical plausibility with clinical real-time requirements. We present a graph-based Ensemble Iterative Estimator that integrates elasticity-inspired regularization into neural network predictions through a mesh-free graph optimization framework to reduce computational complexity (Fig. 9d). The method converts the post-operative facial 3D point cloud $F' = \{f'_j \in \mathbb{R}^3\}_{j=1}^{N_F}$ into a graph structure model $G = (V, E)$ with local neighborhood topological connections. Edge weights are assigned via a Gaussian kernel:

$$\omega_{ij} = \exp(-||\delta_i - \delta_j||^2/(2\sigma_i^2)) \tag{9}$$

to encode both the local stiffness of soft tissues and spatial information within the point cloud.

The displacement field is obtained by solving the constrained system:

$$L_G \boldsymbol{\delta} = 0, \qquad s.t. \boldsymbol{\delta}|_S = \boldsymbol{\delta}_{\text{fix}}, \tag{10}$$

where the constrained point set $S$ consists of spatial points where displacement is fixed to $\boldsymbol{\delta}_{\text{fix}}$ predicted by neural network, and $L_G$ denotes the graph Laplacian operator:

$$(L_G \boldsymbol{\delta})_{ij} = \begin{cases} \sum_{k \in N(i)} \omega_{ik}, & i = j, \\ -\omega_{ij}, & i \neq j \bigwedge j \in N(i), \\ 0, & otherwise. \end{cases} \tag{11}$$

This formulation arises from combining deformation continuity constraints through local neighborhood coupling with neural network-predicted displacement preservation.

When setting $\sigma_i$ as the mean nodal distance $h$, the graph Laplacian operator converges to the continuous Laplacian operator[46]:

$$\frac{1}{h^2}(L_G \delta)_i = \sum_j \omega_{ij}(\delta_i - \delta_j) \overset{h \to 0^+}{\longrightarrow} C\Delta\delta(x_i), \tag{12}$$

which under near-incompressibility conditions of biological tissues predominantly governs shear effects in elastic deformation[47]. This heuristic discretization choice inherently captures shear-dominated deformation patterns through its operator properties, while maintaining mathematical consistency with biomechanical principles.

For efficient computation, we implement a Jacobi relaxation iteration (Supplementary Fig. S7) scheme:

$$\delta_i^{(n+1)} = \frac{1}{(L_G)_{ii}} \left( \sum_{j \in N(i) \cap S} \omega_{ij} \delta_j + \sum_{k \in N(i) \setminus S} \omega_{ik} \delta_k^{(n)} \right), \tag{13}$$

where $N(i)$ denotes the neighborhood of point $x_i$. This iterative process essentially propagates displacements from known to unknown points.

Notably, even when neural network predictions $\boldsymbol{\delta}_{\text{fix}}$ exhibit non-zero deviations from biomechanical priors (resulting in residual terms in the harmonic equation's RHS), the relaxation method effectively reconciles prediction results with mechanical constraints through its robust convergence behavior, thereby achieving rapid high-fidelity facial reconstruction.

The algorithm initializes unknown point weights through neighborhood averaging. Each iteration processes only local neighborhoods with an average size $k$ accounting for 0.05% of the total point cloud $N$, achieving $O(kN)$ complexity. Experimental validation confirms millimeter-scale accuracy within 5 iterations, demonstrates performance advantages in processing $O(10^5)$-scale point clouds. By embedding biomechanical principles into a high-efficiency computational framework, this method establishes a novel paradigm for rapid reconstruction of diverse facial morphologies.

### Combination of multiple loss functions

The network is trained using a combination of loss functions. The first component is a displacement field distance constraint, formulated as the Chamfer Distance (CD) between the predicted postoperative facial point cloud and the ground-truth surface:

$$L_{CD} = \frac{1}{N_F} \left( \sum_{f_i \in F} \min_{f'_j \in F'} \left\| f_i + \Delta f_i - f'_j \right\|^2 + \sum_{f'_j \in F'} \min_{f_i \in F} \left\| f_i + \Delta f_i - f'_j \right\|^2 \right) \tag{14}$$

Additionally, a smoothness regularization term is introduced to promote gradual, spatially coherent deformations:

$$L_{smooth} = \frac{1}{N} \sum_{i=1}^{N} \sum_{j \in N(i)} \left\| \Delta f_i - \Delta f_j \right\| \tag{15}$$

Finally, since only the final displacement field is explicitly supervised, while intermediate states in the LSTM decoding process remain unobserved, we introduce a weak supervision constraint to enforce directional consistency across the iterative steps. Specifically, we encourage the displacement increments at each time step to align with the overall deformation direction. The loss is defined as:

$$L_{prog} = \frac{1}{T} \sum_{t=0}^{T-1} \left\| \sum_{k=0}^{t} \delta^k - \Delta f_{pseudo}^t \right\|^2, \ \Delta f_{pseudo}^t = \frac{t}{T-1} \Delta f \tag{16}$$

### Model evaluation

To comprehensively evaluate the predictive performance of PhysSFI-Net, multiple quantitative metrics were employed. First, for the predicted facial point clouds, the Hausdorff distance (HD) between the ground-truth postoperative facial shapes and the predicted point clouds was calculated to quantify the overall shape accuracy. In addition to this global assessment, a subregional shape error analysis was performed to provide clinically interpretable insights into model performance across anatomically and aesthetically significant facial regions. Specifically, the facial surface was partitioned into five subregions: chin, lips, nose, left cheek, and right cheek.

For each region, the regional Hausdorff distance was computed between the predicted and ground-truth point clouds, enabling a detailed evaluation of spatially localized prediction accuracy and facilitating the identification of anatomically challenging areas.

Second, for reconstructed 3D facial meshes, the average surface deviation error between the ground truth mesh and the reconstructed mesh from the model predictions was computed. These mesh deviation errors were visualized using heatmaps with a color scale ranging from −4.00 mm to +4.00 mm, providing intuitive spatial information about prediction accuracy.

Third, to assess landmark prediction accuracy, we selected 15 clinically significant facial landmarks in the periorbital, midface, perioral, and chin regions. Experienced clinicians independently annotated these landmarks on both the ground truth and the predicted facial meshes, and Euclidean distances between corresponding landmarks were computed within a common coordinate system. Since landmark errors less than 2 mm are typically considered clinically acceptable, we calculated and reported the proportions of landmark prediction errors falling within three ranges: less than 2 mm, between 2 and 4 mm, and greater than 4 mm.

### Multi-center validation

An independent multi-center external validation cohort was established to further assess the robustness and clinical feasibility of the proposed PhysSFI-Net. This cohort consisted of 66 CT scans (33 preoperative-postoperative pairs) from 33 patients with dentofacial deformities who underwent orthodontic-orthognathic treatment. All data were collected from collaborating institutions and processed using the same preprocessing pipeline as the primary dataset, including segmentation, registration, and point cloud generation. The trained model was directly applied to this external cohort for inference.

To comprehensively evaluate clinical applicability, both objective quantitative assessments and subjective clinician-based evaluations were performed. Objective quantitative assessments were performed to measure the geometric accuracy of the predicted postoperative facial morphology. The Hausdorff Distance between the predicted facial point clouds and the corresponding ground-truth postoperative surfaces was calculated to quantify the overall shape discrepancy. A subregional shape error analysis was conducted by partitioning the facial surface into five anatomically significant regions: the chin, lips, nose, left cheek, and right cheek. After reconstructing high-resolution 3D facial meshes from the predicted point clouds, the average surface deviation between the predicted and ground-truth meshes was computed to assess the overall geometric fidelity.

To further investigate the clinical applicability and decision-support value of the proposed model, two complementary clinician-based subjective evaluations were conducted within the external validation cohort: (1) Clinical Acceptability Assessment: Surgeons were presented with the predicted postoperative facial morphology alongside the actual postoperative soft-tissue appearance obtained six months after surgery, displayed in both frontal and lateral views. The predictions were rated using a three-level categorical scale: clinically acceptable, basically acceptable, or inaccurate, reflecting their suitability for surgical planning. (2) Reasoning Consistency Evaluation: To assess agreement with surgeons' clinical expectations, evaluators were provided with the preoperative skeletal structures, detailed surgical plans (including surgical procedures and magnitudes of skeletal movements), and the simulated postoperative skeletal configuration, without access to the actual postoperative facial appearance. The consistency between the predicted outcomes and the anticipated surgical results was rated in both frontal and lateral views as highly consistent, basically consistent, or inconsistent.

Both assessments were independently performed by two experienced orthognathic surgeons. When their evaluations were concordant, the result was accepted as final. In cases of disagreement, a consensus decision was reached through discussion with an additional senior surgeon, with preference given to the more conservative judgment. To minimize potential bias, the reasoning consistency evaluation was conducted prior to the clinical acceptability assessment, which was performed one week later to prevent recall bias and to ensure that prior knowledge of the ground truth did not influence the surgeons' reasoning-based judgments.

### Comparative analysis

To comprehensively evaluate the performance of the proposed PhysSFI-Net, comparative experiments were conducted against multiple representative deep learning models for facial shape prediction. Specifically, we reproduced one of the SOTA models named the Attentive Correspondence-assisted Movement Transformation Network (ACMT-Net) by strictly following the implementation details and settings described in the original publication. This ensured a fair and reliable comparison between PhysSFI-Net and the current leading approach. To evaluate the feature extraction capability of PhysSFI-Net, two baseline models were constructed by adopting DGCNN-based and Point Transformer-based architectures for facial shape prediction. DGCNN captures local geometric relationships through dynamic graph construction and edge convolution operations, whereas the Point Transformer employs a self-attention mechanism to model both local and global contextual dependencies within point clouds[28,29]. These models were compared with PhysSFI-Net using global and subregional shape errors.

Further comparisons between PhysSFI-Net and the SOTA model (ACMT-Net) were performed with respect to mesh-based shape accuracy, including the average surface deviation of the reconstructed three-dimensional facial meshes. Their clinical applicability was evaluated using the external multi-center validation cohort, incorporating both objective quantitative metrics and subjective clinician-based assessments.

### Ablation study

To evaluate the contribution of the sequential deformation modeling component, an ablation study was performed by constructing PhysSFI-Net variants with 0, 1, 2, 3, and 4 LSTM layers. The 0-layer configuration represents a direct deformation prediction framework without sequential modeling. All variants were trained and evaluated under identical experimental conditions.

### Model interpretability analysis

To enhance model interpretability, pGS-CAM (point Grad-Seg Class Activation Mapping) was employed to visualize the contribution of individual points in the preoperative facial point cloud[48,49]. Inspired by Grad-CAM, this method generates saliency maps for each activation layer, highlighting regions that most strongly influence the predicted postoperative morphology. The analysis provides qualitative insights into the skeletal–facial interactions captured by PhysSFI-Net.

### Training details and statistics analysis

We implemented our model using PyTorch library. The network was trained using the Adam optimizer on an NVIDIA A100 GPU with 80 GB of memory. The batch size was set to 8 for all experiments. All statistical analyses were performed by R software (Version 4.1.2). Categorical variables were presented in the form of numbers and percentages, while continuous variables were presented as means ± standard deviations. For the comparison of continuous variables between two groups, the *T*-test was employed for normally distributed continuous variables, and the Wilcoxon rank-sum tests were used for non-normally distributed continuous variables. A *P* value of less than 0.05 was regarded as statistically significant.

## Data availability

The datasets generated and/or analyzed during the current study are not publicly available due to ethical and privacy constraints involving identifiable patient facial data. Source code is available from the corresponding author upon reasonable request (https://github.com/CMFCarter/OGSpredict).

## Code availability

Source code is available from the corresponding author upon reasonable request (https://github.com/CMFCarter/OGSpredict).

## Acknowledgements

This study was supported by National Natural Science Foundation of China (81571022), Shanghai Natural Science Foundation (23ZR1438100) and AI Interdisciplinary Research Program of Shanghai Ninth People's Hospital (JYJC2025010).

## Author contributions

Jiahao Bao: Conceptualization, Data Curation, Methodology, Formal Analysis, Investigation, Visualization, Writing - Original Draft. Huazhen Liu: Data Curation, Methodology, Formal Analysis, Software, Investigation, Validation, Writing - Original Draft. Yu Zhuang, Leran Tao: Data Curation, Methodology, Investigation, Validation, Writing- Reviewing and Editing. Xinyu Xu, Yongtao Shi: Data Curation, Methodology, Visualization, Writing-Reviewing and Editing. Mengjia Cheng, Yiming Wang, Congshuang Ku: Data curation, Investigation, Software, Validation, Writing- Reviewing and Editing. Ting Zeng, Yilang Du, Siyi Chen: Data curation, Validation, Writing-Reviewing and Editing. Shunyao Shen, Suncheng Xiang: Conceptualization, Investigation, Supervision, Resources, Writing- Reviewing and Editing. Hongbo Yu: Conceptualization, Investigation, Supervision, Resources, Project administration, Funding Acquisition, Writing- Reviewing and Editing.

## Competing interests

The authors declare no competing interests.

## Additional information

**Correspondence** and requests for materials should be addressed to Hongbo Yu.